# Attention U-Net as a surrogate model for groundwater prediction

Maria Luisa Taccari[1], Jonathan Nuttall[2], Xiaohui Chen[1], He Wang[1], Bennie Minnema[2] and Peter K. Jimack[1]

[1]School of Civil Engineering, University of Leeds, Leeds, LS2 9JT, UK.
[2]Deltares, Delft, 2629 HV, The Netherlands.

**Abstract**

Numerical simulations of groundwater flow are used to analyze and predict the response of an aquifer system to its change in state by approximating the solution of the fundamental groundwater physical equations. The most used and classical methodologies, such as Finite Difference (FD) and Finite Element (FE) Methods, use iterative solvers which are associated with high computational cost. This study proposes a physics-based convolutional encoder-decoder neural network as a surrogate model to quickly calculate the response of the groundwater system. Holding strong promise in cross-domain mappings, encoder-decoder networks are applicable for learning complex input-output mappings of physical systems. This manuscript presents an Attention U-Net model that attempts to capture the fundamental input-output relations of the groundwater system and generates solutions of hydraulic head in the whole domain given a set of physical parameters and boundary conditions. The model accurately predicts the steady state response of a highly heterogeneous groundwater system given the locations and piezometric head of up to 3 wells as input. The network learns to pay attention only in the relevant parts of the domain and the generated hydraulic head field corresponds to the target samples in great detail. Even relative to coarse finite difference approximations the proposed model is shown to be significantly faster than a comparative state-of-the-art numerical solver, thus providing a base for further development of the presented networks as surrogate models for groundwater prediction.

## 1    Introduction

Groundwater resources are of major importance for residential, industrial and agricultural use. However, the quality and availability of groundwater supplies are significantly affected by their overexploitation around the world, population growth and climate extremes [1]. Consequently, a demanding need exists for quick and accurate evaluation of multiple management alternatives over long time horizons. The last 30 years have seen the development of several physics-based numerical models for simulating groundwater systems, with Finite Difference (FD) and Finite Element (FE) discretizations of the partial differential equations (PDEs) as the most used and classical methodologies [2, 3]. These techniques calculate the hydraulic head by iteratively solving an implicit system of equations at each time step in the discretized time and flow domains. Running the groundwater model in a complex system within a large domain and with reasonable accuracy incurs numerical challenges and an excessive computational demand [4, 5]. As the computational cost increases super-linearly with the number of unknowns in the discretization, long runtimes are a major challenge when high resolution is required or when many executions are necessary, such as in uncertainty analysis, sensitivity analysis, and inverse modelling. Mens et al., [6], discuss the case of the National Water Model (NWM) that is used for national policy-making on drought risk management in the Netherlands and whose heavy computational burden poses limits to quickly responding to policy questions. The authors advocate the need for a fast simple model that describes all relevant processes and is quick enough to explore many scenario and strategy combinations for long time series. Furthermore, groundwater flow simulations require the reconstruction of subsurface heterogeneities and the physical properties of the aquifer as inputs to the model, for which only limited direct observations are available. Inverse modelling is used to estimate the unknown parameters of the system, taking into account their stochasticity.

Traditional approaches to the inversion problem correspond to iterative techniques and necessitate a large number of forward model runs. As the number of unknown parameters increases, forward operations become extremely computationally demanding. Surrogate models are cheaper-to-run models which approximate the response of a complex and computationally intensive model. Surrogate models have been used in a number of groundwater studies, such as for optimization design [7, 8] and uncertainty quantification problems[9-11], to name a few. Reduced-fidelity models simplify the level of complexity of the physical processes of the full-order model, e.g. by projecting the governing equations into a transformed space of smaller dimension. Projection-based techniques can accurately retain the underlying structure of the full-order model; however these methods can suffer stability and robustness issues [12, 13], they are highly code-intrusive and they cannot efficiently treat strong nonlinearity [14]. Data-driven models learn the response of the system from the simulation data in a supervised manner. Gaussian processes have been successfully applied to uncertainty quantification tasks for which the training data are limited but they rely on specific *a priori* assumptions on the relationship between the input and the outputs and have high computational costs when dealing with large datasets. [15, 16]

Deep neural networks are universal function approximators and are becoming increasingly common surrogate models for solving problems within the fields of physics and engineering. These techniques have been applied for solving PDEs in high-dimensional settings and nonlinear systems, with potential applications in parameter estimation and uncertainty quantification. The reader is referred to Karniadakis et al. [17] for a review on the strengths, limitations, current applications and outlook of this class of deep learning algorithms.

Recently interest has grown for learning complex nonlinear, multiscale, and high dimensional mappings of subsurface processes. In the work of Geneva and Zabaras [18], convolutional neural networks (CNNs) for physics-constrained learning show exceptional performance, with solutions obtained an order of magnitude faster than with state-of-the-art numerical solvers. They train deep auto-regressive convolutional neural network models to learn the dynamics of three transient PDEs (1D Kuramoto-Sivashinsky equation, 1D Burgers' equation and the 2D coupled Burgers' system) without any off-line training data. Several studies adopted an adversarial network framework for surrogate methods for a single-phase flow forward model and a multiphase flow forward model [19, 20]. Dagasan et al., [21], argue that the use of a conditional generative adversarial network (cGAN) as a surrogate forward model for groundwater systems can reduce the computational time by up to 80% compared to the numerical solver MODFLOW.

Deep neural networks were chosen in this study largely due to their scalability and their ability to learn based on a few *a priori* assumptions. The first refers to the capacity to learn from massive amounts of data. Compared to a Gaussian process, whose runtime scales poorly with the size of the datasets, deep neural networks can assimilate large amounts of multi-fidelity observational data, even in partly understood, uncertain and high-dimensional contexts. Compared to reduced order model techniques, which aim to bring the physical relationships of full order models at a much lower dimension, deep neural networks do not assume any prior assumption that constrains the relationship between input and output samples. This flexibility can lead deep neural networks to learn complex relationships, thus increasing their modelling power but at the cost of a lower interpretability.

The encoder-decoder architecture consists of a contracting and an expansive path. It shows robust and accurate performance in various tasks including machine translation problems [22], semantic segmentation [23] or depth regression [24]. Initially developed for biomedical image segmentation [25], U-Net is an encoder-decoder network which uses fully convolutional networks and requires highly limited training samples. U-net based architectures have been applied across a wide spectrum of application areas, such as image super resolution, style transfer, text-to-image translation and image-to-image translation [26]. Mo et al.,[27], developed a deep convolutional encoder-decoder network method as a surrogate model for transient multiphase flow models. Given the large approximation errors in the concentration fields near the source release location, the authors assign an additional weight to the loss at the eight pixels around the source release location in order to improve the surrogate predictive capability. Attention models address this limitation by allowing the model to learn to focus selectively on the relevant parts of the input. Attention has recently become an essential component of neural architectures within diverse application domains [28-30]. Attention U-net makes use of attention gates in order to focus on specific parts of the image that are of importance while paying little attention to unnecessary areas [23, 31].

The purpose of this paper is to propose an Attention U-Net network as a surrogate model for the forward operator in groundwater modelling. The encoder-decoder model learns the mapping between model inputs and output for deterministic, steady-state solutions of the two-dimensional groundwater flow equation given a highly heterogeneous subsurface domain. The surrogate model accurately captures the nonlinear relationship between the hydraulic conductivity and the subsurface groundwater map. The model dynamically pays attention to only the parts of the input where flow can take place in a manner that helps the network in learning the mapping effectively.

The rest of the paper is organized as follows. Section 2 presents the adopted image-to-image deep learning approach and the architecture of the Attention U-Net employed. Section 3 provides an overview of the problem formulation and model set-up along with training of the surrogate model. The proposed method is evaluated with and without attention gates in section 4. Finally, the conclusions are formulated in the last section.

## 2 Methodology

### 2.1 Surrogate Modelling as Image-to-Image Regression

A surrogate model $\hat{f}(x, \theta) \approx y$ approximates the 'ground-truth' function $y = f(x)$ where $f: X \to Y$ is the mapping between the input domain X and the output domain Y, $x \in X$ is the input, $y \in Y$ is the output and $\theta$ are the model parameters. In the case of forward solving of PDEs with machine learning, the ground truth mapping represents some combination of the solution of the PDEs governing the physical system, and the surrogate model $\hat{y} = \hat{f}(x, \theta)$ is trained using a dataset D of N simulation data: $D = \{x^i, y^i\}_{i=1}^N$.

By adopting an image-to-image regression approach, the surrogate modelling can be treated as an image regression problem. By solving the PDE over a spatial domain, such as 2D regular grids, the simulation data can be thought as images, with inputs $x^i \in \mathbb{R}^{d_x \times H \times W}$ and outputs $y^i \in \mathbb{R}^{d_y \times H \times W}$ where $d_x$ and $d_y$ are the number of input and output images with a resolution of H × W (height × width). The surrogate modelling problem becomes an image-to-image regression problem with the regression function $\hat{f}: \mathbb{R}^{d_x \times H \times W} \to \mathbb{R}^{d_y \times H \times W}$ [32].

## 2.2 Encoder-decoder model

Encoder-decoder is a learning method with an analysis path (encoder) and a synthesis path (decoder). The encoder network transforms high-dimensional unlabeled input data x into low-dimensional embeddings z (latent space) and the decoder maps z to the intended output y = decoder ∘ encoder(x). The input is passed through a series of layers that progressively down sample until a bottleneck layer, at which point the decoder restores the spatial dimensions to produce the output images. Intuitively, the model corresponds to a coarse-refine process: the encoder reduces the spatial dimension of the input image to high-level coarse features, and the decoder recovers the spatial dimension by refining the coarse features. The assumption is that the input and output images share the underlying structure, or they are different renderings of the same underlying structure, that is their structures are roughly aligned [26].

As the goal of this study is to generate a targeted output image corresponding to given inputs, the Encoder-Decoder model learns the mapping x → y from a conditioning input image x to the output image y. The network converts images from the source to target domains, where the first corresponds to the initial, boundary conditions and model parameters and the latter to the resolution of the governing equation given those constrains.

## 2.3 Deep Convolutional Neural Networks

CNNs [33, 34] are popular deep learning networks specialized in image processing [35, 36]. While the first layers detect basic features, deeper convolutional layers learn higher representation. A convolution layer is a linear transformation that highlights the presence of a given feature in the map while preserving spatial information in the input image [37]. Given a 2-D input image and a square kernel ω with size m, the convolutional layer outputs the value at location (i,j) by summing up the contributions from the previous layer cells $y^{l-1}$ weighted by the filter components; then, the nonlinearity σ is applied.

$$y_{ij}^l = \sigma \left( \sum_{a=0}^{m} \sum_{b=0}^{m} w_{ab} y_{(i+a)(j+b)}^{l-1} \right) \qquad (1)$$

The stride of the convolutional layer is a parameter that determines the number of pixel shifts between two successive moves of the filter, while the padding indicates the amount of pixels with value zero added at each side of the boundaries of the input. The rectified linear unit function (ReLU) is a piecewise linear function that outputs the input if it is positive and zero if negative. The Leaky ReLU with slope coefficient α modifies the function to allow a small, negative, output when the input is negative:

$$\sigma(x) = \begin{cases} x & if\ x > 0 \\ \alpha x, & otherwise \end{cases} \qquad (2)$$

Batch normalization and dropouts are used to stabilize training and mitigate overfitting [38]. A dropout layer selects a random set of units from the preceding layer and ignores their output, while batch normalization standardizes the layer's inputs by calculating the mean and standard deviation across the batch.

## 3  Application

### 3.1  Groundwater model and datasets

Consider steady-state groundwater flow in saturated media satisfying the fundamental governing equation [39]:

$$\nabla \cdot (\boldsymbol{K}\nabla h) + q = 0 \qquad (3)$$

The piezometric head h [L] is the field variable of interest, $\boldsymbol{K}$ is the input hydraulic conductivity [L/T] and q represents the source (or sink) terms [$L^3 T^{-1}$].

The problem of this study consists of steady-state flow in a single-layer model representing a heterogeneous confined aquifer. Initially, in this work, only Dirichlet boundary conditions are considered and the groundwater head values are fixed in the cells in which the allocated head is known. The model takes in an input image with three channels: head values, boundary markers and spatially varying hydraulic conductivity (Figure 1). Dirichlet boundary conditions are imposed on the four sides of the square domain. Head is constant at up to three random locations across the domain, representing wells. The source term q is set to zero. The second channel of the input image is a binary mask where the boundary markers identify the cells with a fixed value, i.e. well locations and boundary cells as defined by the first source image. The last input channel defines the heterogeneous media. The conductivity field $\boldsymbol{K}$ of the highly-heterogeneous aquifer is a Gaussian random field [40] in which the values of hydraulic conductivity are taken from a finite set of values.

This application example demonstrates the capability of an Attention U-Net to successfully learn and simulate a common hydrologic situation using an image-to-image translation approach. The model is trained to predict the output fields consisting of the spatial components of the groundwater head in the domain. These predictions are compared against simulation results obtained by the fully-implicit finite difference model MODFLOW [2], here called "target images", bearing in mind that finite difference results provide an approximation of the partial differential equation and are not error free.

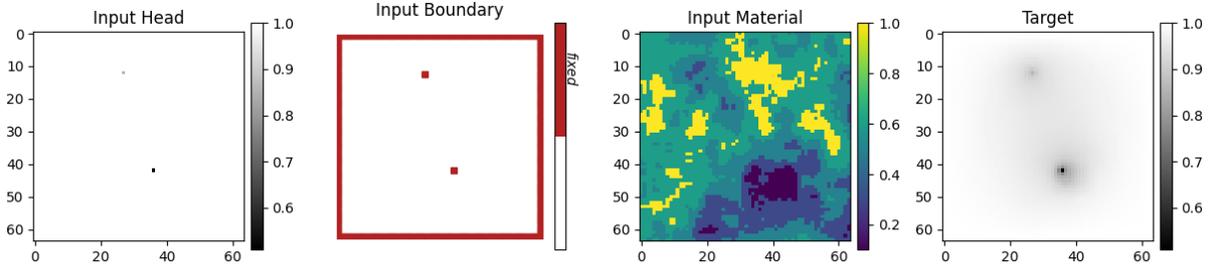

*Figure 1: Input and output channels for an example taken from the training dataset. (Left to Right) The values of the piezometric head at the boundaries (input- channel #1); the location of the boundaries (input- channel #2); the hydraulic conductivity field (input- channel #3); the hydraulic head in the whole domain (output).*

## 3.2 Network architecture

The Encoder-Decoder model is implemented as Attention U-Net [23, 25] and the employed network architecture is shown in Figure 2 (for the case of a 64x64 input image, as used in our computational tests). This is an encoder-decoder model with skip connections. In the down-sampling half the inputs are encoded with a series of CNNs with kernels of size 4, stride of 2 and padding set to 1. In each block, CNN is followed by a Batch Normalization layer, Dropout with rate 0.5 and a Leaky ReLU with slope 0.3. As the number of filters increases to 512 and the size of the input images reduces to 4x4, the encoder captures high-level abstract information. In the up-sampling half the representations are expanded spatially and the number of channels is reduced by a series of CNNs and up-sampling layers. The last up-sampling layer is followed by a transposed convolution layer with a sigmoid activation function to ensure predicted values between 0 and 1. Skip connections link the layers in the encoder with corresponding layers with the same-sized feature map in the decoder [25]. The only difference between Attention U-Net and the original U-Net architecture is that in the Attention U-Net network skip connections are additionally passed through attention gates, which use additive soft attention [23]. The attention coefficients are larger if the vector from the next lowest layer of the network in the up-sampling path and the corresponding vector from the encoder going through the skip connection are aligned. The weights are multiplied element-wise to the original vector which passes along in the skip connection. In this way, the attention gate (AG) mechanism allows the U-Net to suppress irrelevant regions and focus more on target structures of varying size and shape. For reproducibility full details of the network architecture are described in Appendix A.

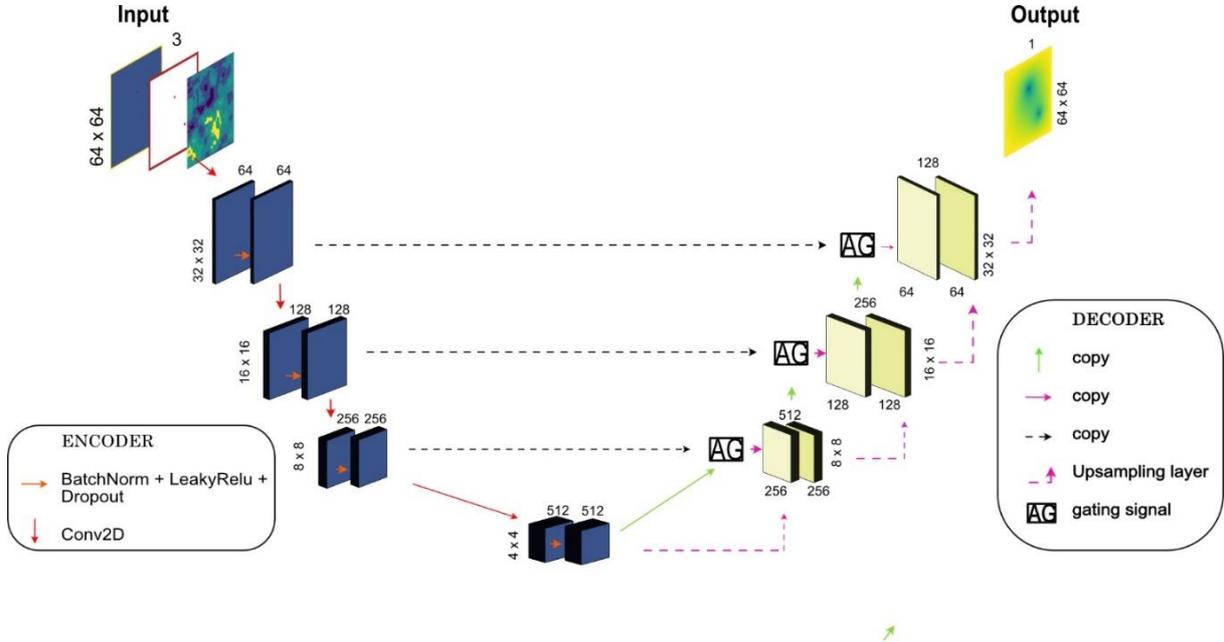

*Figure 2: Attention U-net architecture as the surrogate model. The model has three input channels and one output channel as shown illustrated in Figure 1. The first part is the down-sampling half (left, in blue): the inputs are encoded with a series of CNNs with kernels 4x4 and stride of 2 and down-sampling layers. The second half (right, in yellow) is the up-sampling part: the representations are expanded spatially and the number of channels is reduced.*

## 3.3 Loss function

The aim of the regression task is to minimize the mean square error (MSE) between the generated samples and training data. The network computes the average loss across a mini-batch of size $N_b$:

$$\mathcal{L}_{MSE} = \frac{1}{N_b} \sum_{j=1}^{N_b} \frac{\sum_{i=1}^{n}(y_{j,i} - \hat{y}_{j,i})^2}{n} \quad (4)$$

where y is the training image, $\hat{y}$ is the image generated by the network and n denote the total number of pixels of each image. The networks tries to be near the ground truth output in an L2 sense.

## 3.4 Network Training

The model is trained in supervised fashion. For the examples described in the following section a dataset consisting of 32000 training data is used, and the development of the loss function is compared with 8000 validation data. The size of the training dataset is big enough to ensure the model's ability to generalize and the generation time is less than 3 hours.

The losses are minimized using the Adam optimizer [41] with a starting learning rate $\alpha = 8 \times 10^{-4}$. The network is trained for 130 epochs. Training converged after approximately 2 hours, by training the models on an Intel(R) Xeon(R) GPU Tesla K80.

The quality of the trained network is evaluated by reporting the coefficient of determination ($R^2$) and the root mean squared error (RMSE) between each pixel value from the target image and each pixel value from the generated image:

$$R^2 = 1 - \frac{\sum_{i=1}^{N} \|y_i - \hat{y}_i\|_2^2}{\sum_{i=1}^{N} \|y_i - \bar{y}\|_2^2}, \quad RMSE = \sqrt{\frac{1}{N}\sum_{i=1}^{N}\|y_i - \hat{y}_i\|_2^2} \quad (5)$$

where y is the target image, $\bar{y}$ is the mean of the target images of the dataset $\bar{y} = \frac{\sum_{i=1}^{N} y_i}{N}$, $\hat{y}$ is the network prediction, and N is the total number of samples. The two selected metrics between them yield complementary and representative information for the evaluation of the trained model.

## 4 Results and Discussion

### 4.1 Model predictions

This section presents both qualitative and quantitative techniques to test the performance of the model. The test case considers a square domain $\Omega = [0, 64] \times [0, 64]$ consisting of 64 rows and 64 columns, with the width of each cell equal to one. The boundary is assumed to be constant head boundary with head of 1, while the imposed head values in the wells lie in the range [0.5,1]. The number of wells, their locations and their values are randomly selected and vary for each data sample. The conductivity field, K, of the highly-heterogeneous aquifer is generated as a continuous Gaussian random field, which is then discretized into a finite set of values. The heterogeneous hydraulic conductivity field has values belonging to 5 different classes (0.1, 0.325, 0.55, 0.775, 1.) which can be thought of as 5 soil types distributed within the model. The model is trained with loss function in Eq. 4 and the target response is the finite difference simulation.

To illustrate the superior performance of the proposed Attention U-Net architecture against the original U-Net network architecture, the U-Net network without attention gates is also trained using the same training sets and parameters. At the end of the training, the Attention U-Net network achieves a RMSE of $1.98 \times 10^{-3}$ and a $R^2$ score of 0.996, while those obtained by U-Net are 3.78 × 10$^{-3}$ and 0.986, respectively (Figure 3).

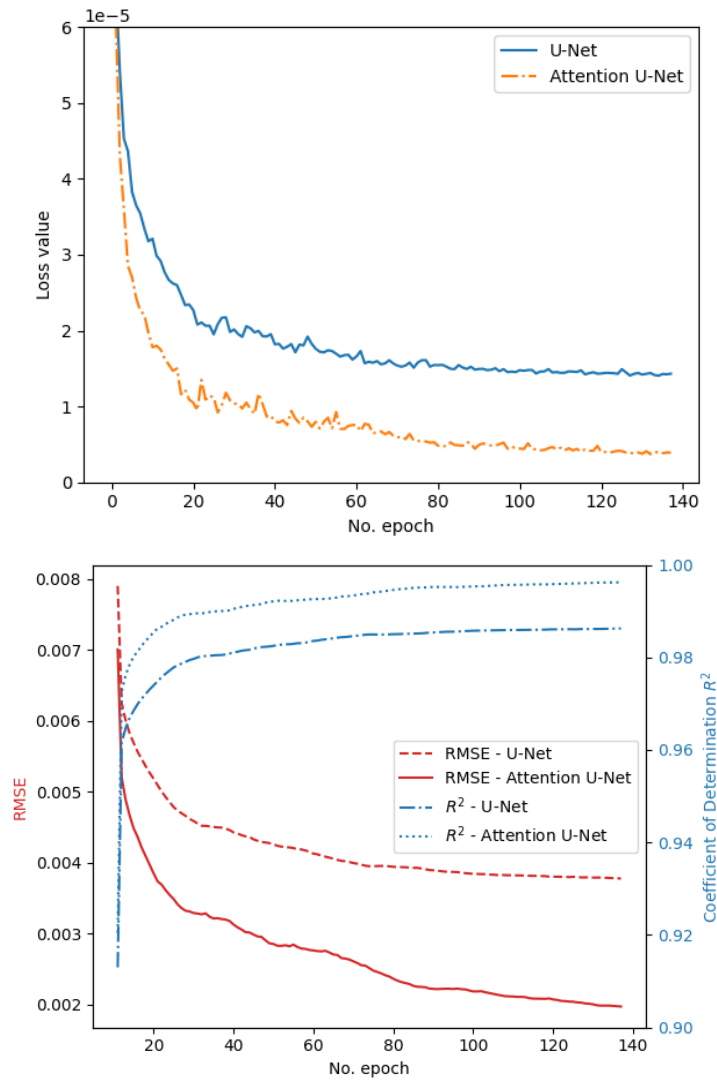

*Figure 3 (Top to Bottom) Loss curves for U-Net (solid line) and Attention U-Net (dashed line); RMSE and R$^2$ scores of the model evaluated on the training dataset for U-Net and Attention U-Net.*

Figure 4 provides a comparison of generated images of groundwater head with the target images for 5 random examples taken from the test dataset with a set size of 4000 samples. The Attention U-Net model has learnt to map the flow patterns: it generates accurate predictions for varied input samples that are unseen during training. The predictions match the target images very well: the model predicts the correct value of groundwater head and the pattern of its distribution. The model is able to identify and focus on salient image regions: the attention coefficients are highest at the boundary of the domain and near the well locations, while they are low in the areas with small head distribution gradients.

When trained without attention gates, U-Net can predict the values of the groundwater head in the domain, but the generated outputs have some minor deviations especially at a distance from the source area and the head gradients are smaller. The use of the attention mechanism significantly improves the accuracy of the results.

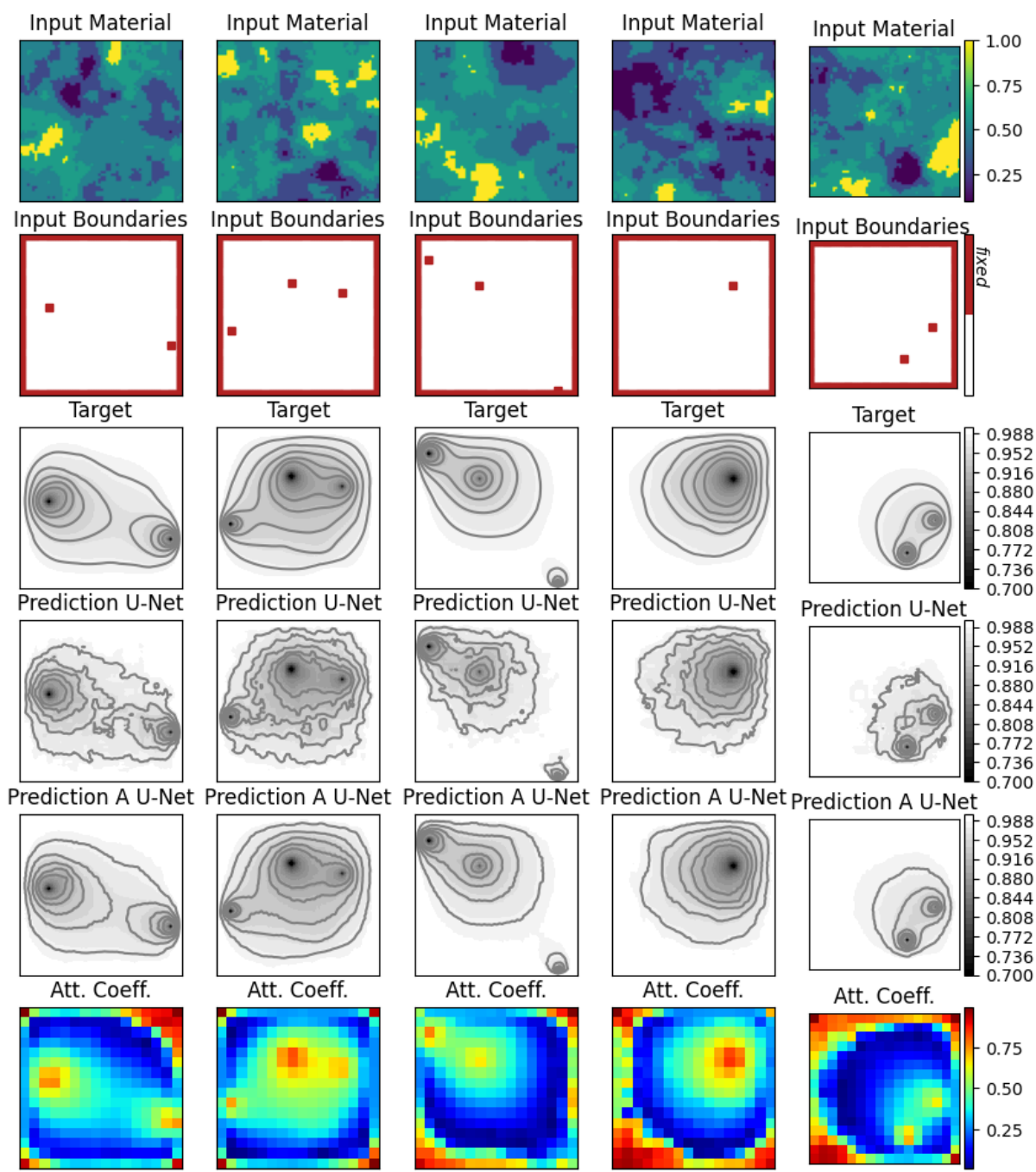

*Figure 4 Comparison between the target sample (MODFLOW) and the learned solution (prediction) for the U-Net and the Attention U-Net models for five randomly selected samples from test dataset. (From top to bottom) The input spatially varying hydraulic conductivity; the location of the input boundaries; the target prediction; the result for the U-Net; the result for the Attention U-Net (A U-Net) ; the attention coefficients learnt by the Attention U-Net. The images in each row share the same colour map with the values given in the rightmost column. The contour lines in the target and prediction images represent the values: 0.9, 0.92, 0.94, 0.96, 0.98.*

Figure 5 visualizes the attention coefficients obtained from two test images with respect to training epochs. During the first 20 epochs, the loss function rapidly decreases (Figure 3, top) and the attention gates learn to identify the location of the wells, the boundaries and a rough outline of the area with large head distribution gradients. By training the network for longer epochs, the attention coefficients are gradually updated and refined to focus on areas with large head distribution gradients.

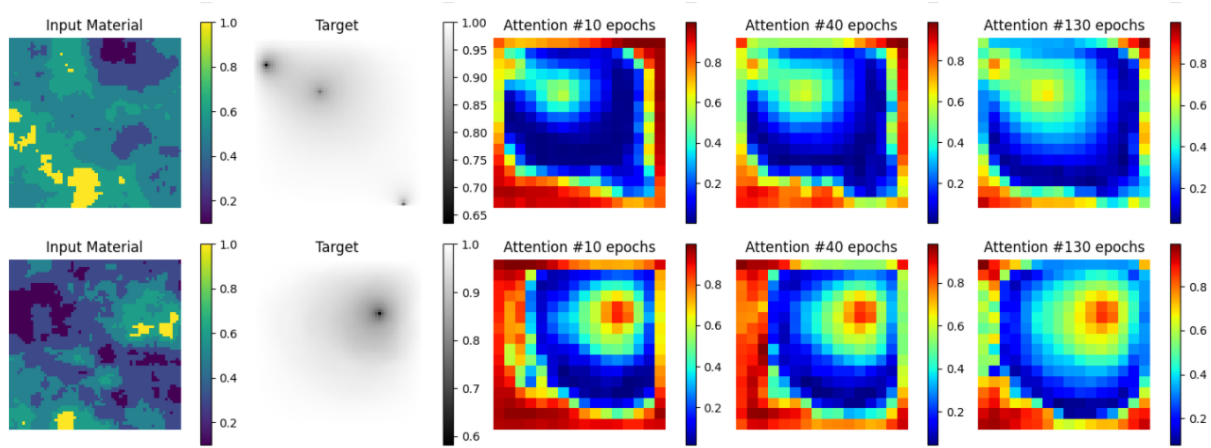

*Figure 5 Example of attention coefficients learnt by the Attention U-Net network across different training epochs (10, 40, 130) for two random samples of the test dataset.*

Appendix B shows the 5 predictions with highest and lowest mean square error between the generated samples and training data out of 500 random samples from the test dataset. The errors are localized near the wells and the difference between the generated and target images is almost negligible even for the samples with the highest error.

## 4.2 Model evaluation

To test the performance of the model, its computational power is compared with the MODFLOW engine. Table 1 presents the processing time required for running the forward operators averaged on 10 examples. In order to have a fair comparison between the two, the tests are performed on the same hardware. The CPU used is Intel(R) Xeon(R) CPU @ 2.20GHz and the GPU is Tesla K80. The results demonstrate that Attention U-Net requires less computational power than MODFLOW. This experiment reveals a 75% computational reduction for the data-driven model, showing its capability to be used in forward simulations with less computational demand than the state-of-the-art numerical solver. When applying the method to computationally more expensive forward models, such as in large-scale non-linear system, the computational cost of the neural network will remain low and significant computational savings can be expected.

*Table 1 Wall-clock time for finite difference and Attention U-net obtained by averaging 10 independent simulation run times.*

|  | **Hardware** | **Backend** | **Wall-clock time(s)** |
|---|---|---|---|
| **Finite Difference** | Intel(R) Xeon(R) CPU @ 2.20GHz , GPU Tesla K80 | MODFLOW FloPy | 0.184 |

| Attention U-Net | Intel(R) Xeon(R) CPU @ 2.20GHz, GPU Tesla K80 | Tensorflow | 0.046 |

Dropout at inference time can be considered equivalent to Bayesian approximation in deep Gaussian processes and the neural network uncertainty can be quantified following the approach proposed by Gal and Ghahramani [42]. At test time, the same input is passed 1000 times to the network with random dropout; the mean and the standard deviation of the generated images give an estimation of the prediction interval. Figure 6 presents the results for 3 random samples: the uncertainty is null at the boundaries and highest in the vicinity of the wells, which is also the region with highest errors. Compared to the finite difference solver, whose response is deterministic, this method allows one to estimate the uncertainty of the model.

The generalization capabilities of the network are presented in Appendix C. The model is able to extrapolate to out-of-distribution inputs, especially for different values of hydraulic conductivity and less so for increasing number of wells.

It is worth pointing out that the effect of using attention gates on the uncertainty and generalization capabilities of the model hasn't been addressed in the current study. Future work should investigate this relation and consequently explore how generalization on out-of-distribution input samples can be improved.

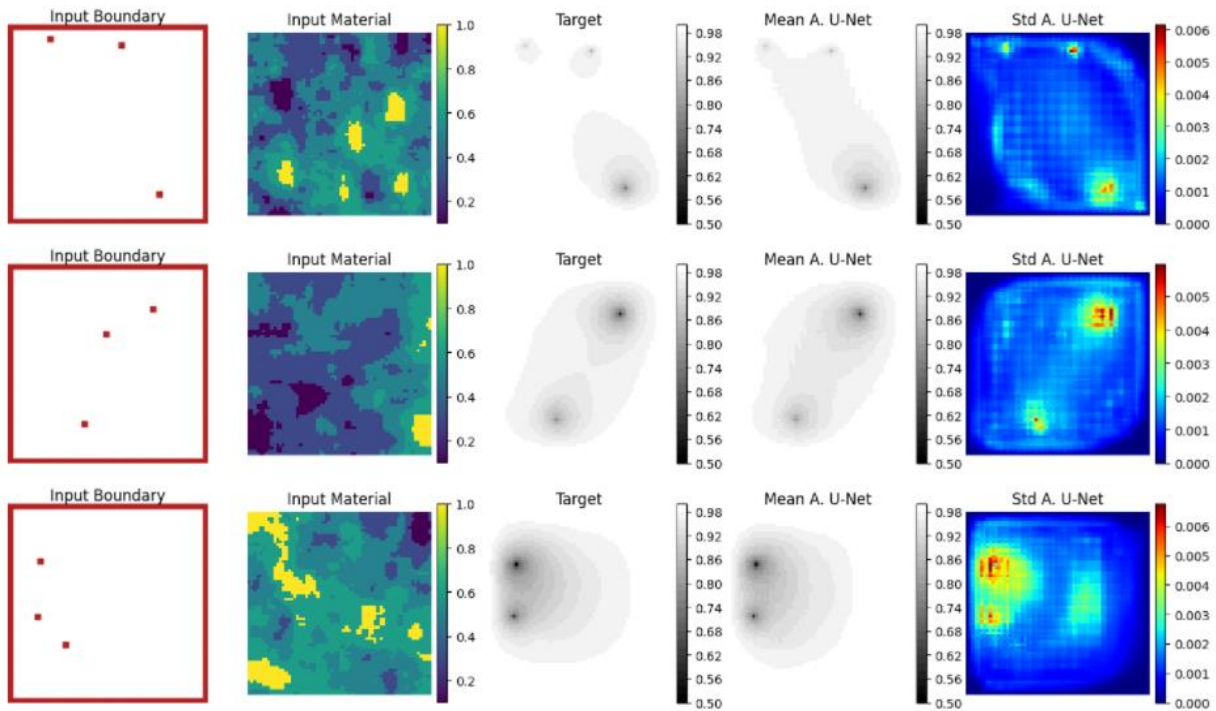

*Figure 6 Model uncertainty: estimate of output mean and standard deviation of the surrogate model for three randomly selected samples from test dataset. From left to right: the location of the input boundaries; the input spatially varying hydraulic conductivity;*

*simulated output obtained with MODFLOW; estimate mean of 1,000 predicted output with the Attention U-net surrogate; estimate of output standard deviation obtained with the data-driven surrogate model.*

## 5    Conclusion and Future Work

This paper presents a convolutional encoder-decoder network to quickly calculate the steady-state response of a groundwater system. The data-driven surrogate model is trained and tested in different scenarios in which the groundwater head values in the whole domain need to be inferred from the hydraulic head at the locations of the wells. The square domain is a Gaussian random field with a spatially varying hydraulic conductivity. When trained by minimizing the departure from the target images, the proposed U-Net model easily learns the nonlinear relation between inputs (hydraulic conductivity fields and boundary conditions) and output (the hydraulic head field).

A significant contribution of the proposed framework is to incorporate attention gates, which allow the network to identify and focus on the salient regions of the image. The visualization of the attention coefficients demonstrate that the model has learnt to pay attention to areas with large head distribution gradients. The attention mechanism improves the network's approximation accuracy and reduces the model uncertainty. The application of the data-driven surrogate method in solving forward simulations gives very accurate results but requiring much less computational time than the state-of-the-art numerical solver.

One attractive property of this methodology is that the learning is carried out offline. Training converged after less than 3 hours on an Intel(R) Xeon(R) GPU Tesla K80, which can be considered as a low training time compared to typical deep learning models. Once the model is trained, its weights and parameters do not need to be further tuned. The choice of the hyperparameters and the specificities of the U-Net architecture have been chosen based upon manual variation (as opposed to systematic optimization) to give accurate results with low computational time with little apparent sensitivity. Future work could include a more robust hyperparameter tuning study, with a quantitative sensitivity analysis.

In the current study, only Dirichlet boundary conditions were applied to the borders of the domain and the locations of the wells. An additional natural extension of our work is to investigate how well the model generalizes to different and mixed types of boundary conditions. Discretization is another important factor to consider. The present work has been limited to data samples with the same resolution. Many questions remain open related to the discretization of the sample data: e.g. the generalization of the trained model to different discretizations and the amount of training data required if the model needs to be retrained for different resolutions.

The authors plan to further develop the presented model for more complex, larger and uncertain systems. This could include time dependent problems, three dimensional simulations and coupled transport through porous media – all of which are likely to require larger training data sets and potentially deeper networks. Another potential extension is the incorporation of prior information directly into the learning process by imposing a physics constraint in the loss function. Physics-informed learning could increase the speed of inference while requiring less data for the training process. Finally, in this study the network has been trained using synthetic data but the potential use of the proposed model holds promises for the solution of practical applications due to its data-driven nature.

**Acknowledgments**

This work was supported by the Leeds-York-Hull Natural Environment Research Council (NERC) Doctoral Training Partnership (DTP) Panorama under grant NE/S007458/1 and by Rijkswaterstaat (Geert Menting) in the project "Fysica-gebaseerde neurale netwerken in grondwater" under KPP BO06 2018 and KPP BO06 2019. The authors acknowledge the additional guidance and practical expertise provided by the Dutch research institute Deltares.

**Appendix A    Network architecture**

This appendix discusses details in the models used. Both U-Net and Attention U-Net have 4 downsampling layers and 4 upsampling layers (Table A.1). Each layer consists of a series of CNNs with kernels of size 4, stride of 2 and padding set to 1, followed by a Batch Normalization layer and Dropout with rate 0.50. The nonlinear activation is Leaky ReLU with slope 0.3 for the downsampling layers and ReLU for the upsampling ones (Table A.2 and Table A.3). Skip connections concatenate the layers in the encoder with corresponding layers in the decoder. The network of Attention U-Net additionally has attention gates which are implemented as according to the work of Oktay et al. [23]. The total number of parameters of the network is $1.36 \times 10^7$, of which $7.35 \times 10^6$ are for the attention gates.

*Table A.1 Network architecture: internal layers, input and output feature maps and number of parameters.*

| Layers | Input Shape | Output Shape | Parameters |
|---|---|---|---|
| Input Layer | (64, 64, 3) | (64, 64, 3) | 0 |
| Downsampling* | (64, 64, 3) | (32, 32, 64) | 3072 |
| Downsampling | (32, 32, 64) | (16, 16, 128) | 131584 |
| Downsampling | (16, 16, 128) | (8, 8, 256) | 525312 |
| Downsampling | (8, 8, 256) | (4, 4, 512) | 2099200 |
| Attention Gate | [(4, 4, 1024), (8, 8, 256)] | (8, 8, 256) | 3412481 |
| Upsampling | (4, 4, 1024) | (8, 8, 256) | 2164992 |
| Concatenate - Skip Connection | [(8, 8, 256), (8, 8, 256)] | (8, 8, 512) | 0 |
| Attention Gate | [(8, 8, 512), (16, 16, 128)] | (16, 16, 128) | 3150337 |
| Upsampling** | (8, 8, 512) | (16, 16, 128) | 1066112 |
| Concatenate - Skip Connection | [(16, 16, 128), (16, 16, 128)] | (16, 16, 256) | 0 |
| Attention Gate | [(16, 16, 256), (32, 32, 64)] | (32, 32, 64) | 788737 |
| Upsampling** | (16, 16, 256) | (32, 32, 64) | 266816 |
| Concatenate - Skip Connection | [(32, 32, 64), (32, 32, 64)] | (32, 32, 128) | 0 |
| Conv2DTranspose | (32, 32, 128) | (64, 64, 1) | 2049 |

*without Batch Normalization and without Dropout
**without Dropout

| | |
|---|---|
| Total parameters: | 13,610,692 |
| Trainable parameters: | 13,604,548 |
| Non-trainable parameters: | 6,144 |

*Table A.2 Second downsampling layer with input (32, 32, 64) with 128 filters of size 4x4, stride equal to 2 and zero padding.*

| Layers | Input Shape | Output Shape |
|---|---|---|
| Conv2D | ( 32, 32, 64 ) | ( 16, 16, 128 ) |
| Batch Normalization | ( 16, 16, 128 ) | ( 16, 16, 128 ) |
| Dropout | ( 16, 16, 128 ) | ( 16, 16, 128 ) |
| LeakyReLU | ( 16, 16, 128 ) | ( 16, 16, 128 ) |

*Table A.3 Last upsampling layer with input (16, 16, 256) with 64 filters of size 4x4, stride equal to 2 and padding set equal to 1.*

| Layers | Input Shape | Output Shape |
|---|---|---|
| UpSampling2D | ( 16, 16, 256 ) | ( 64, 64, 256 ) |
| Conv2D | ( 64, 64, 256 ) | ( 32, 32, 64 ) |
| Batch Normalization | ( 32, 32, 64 ) | ( 32, 32, 64 ) |
| Dropout | ( 32, 32, 64 ) | ( 32, 32, 64 ) |
| ReLU | ( 32, 32, 64 ) | ( 32, 32, 64 ) |

**Appendix B    Worst and best Model predictions**

This appendix shows the 5 predictions with highest and lowest the mean square error out of 500 random samples from the test dataset. The samples with highest errors present multiple wells with wide plumes which cover most of the domains (Figure B.1); on the contrary, the best predictions are those in which the salient region is limited (Figure B.2). In all cases, highest errors are localized near the wells. It is worth noticing that even when the error is higher, the MSE is in the order of $10^{-5}$ and the difference between the generated and target images is almost negligible.

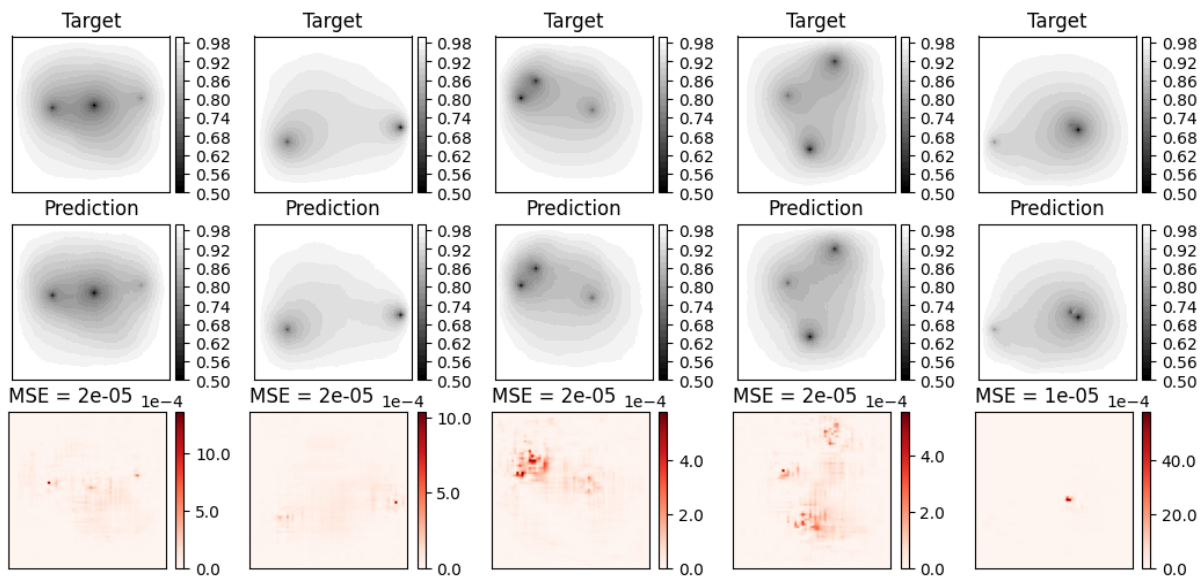

*Figure B.1 Worst 5 predictions out of 500 test samples (highest MSE values)*

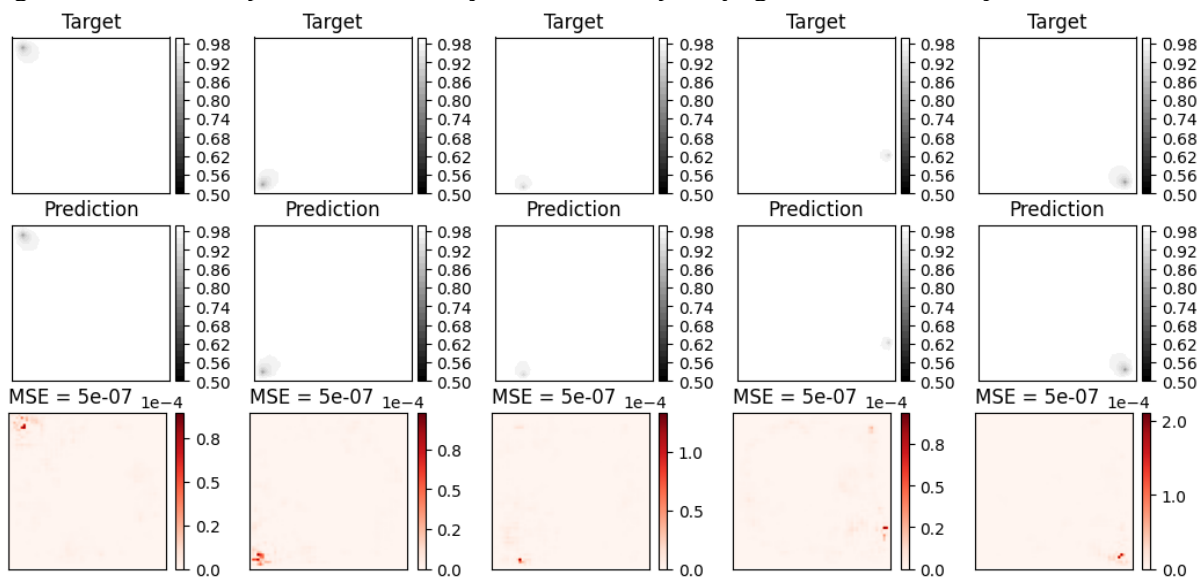

*Figure B.2 Best 5 predictions out of 500 test samples (lowest MSE values)*

**Appendix C    Generalization**

So far the model has been both trained and tested for different scenarios in which the groundwater head values in the whole domain is inferred from the piezometric head at the locations of up to three wells and the spatially varying hydraulic conductivity has values belonging to 5 different classes between 0 and 1. Here we consider testing the model on cases that have different numbers of well locations and different values for the hydraulic conductivity between 0 and 1. Figure C.1 shows the MSE error for the model tested on four new input distributions. The figure shows that the model is able to generalize well given different values of the hydraulic conductivity, but less so for increasing number of wells.

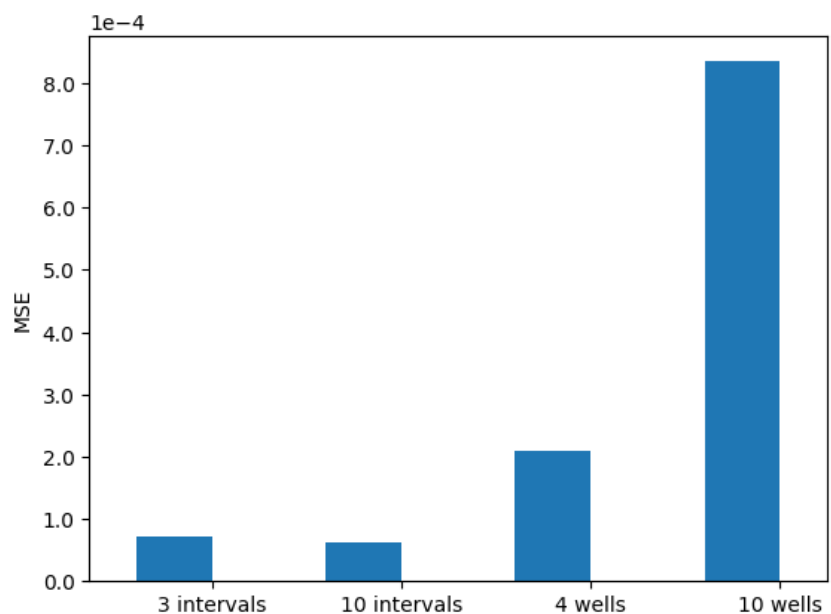

*Figure C.1 Generalization to new input distributions. MSE error when the model is evaluated with hydraulic conductivity having 3 values; hydraulic conductivity values belonging to 10 intervals; 3 wells in random locations the domain; 10 wells in random locations the domain. Each test set contains 1000 samples.*